\documentclass[a4paper,fleqn]{cas-sc}
\usepackage{xcolor}
\usepackage[authoryear]{natbib}
\usepackage{lineno}
\def\tsc#1{\csdef{#1}{\textsc{\lowercase{#1}}\xspace}}
\tsc{WGM}
\tsc{QE}
\begin{document}
\let\WriteBookmarks\relax
\def\floatpagepagefraction{1}
\def\textpagefraction{.001}

% Short title
\shorttitle{}    

% Short author
\shortauthors{}  

% Main title of the paper
\title [mode = title]{Solving the enigma: Enhancing faithfulness and comprehensibility in explanations of deep networks}  

% Title footnote mark
% eg: \tnotemark[1]
%\tnotemark[1] 

% Title footnote 1.
% eg: \tnotetext[1]{Title footnote text}
%\tnotetext[1]{} 

% First author
%
% Options: Use if required
% eg: \author[1,3]{Author Name}[type=editor,
%       style=chinese,
%       auid=000,
%       bioid=1,
%       prefix=Sir,
%       orcid=0000-0000-0000-0000,
%       facebook=<facebook id>,
%       twitter=<twitter id>,
%       linkedin=<linkedin id>,
%       gplus=<gplus id>]
\author[1,2]{Michail Mamalakis}%[<options>]
\cormark[1]
%\fnmark[1,2]
\ead{mm2703@cam.ac.uk}
\author[3,4]{Antonios Mamalakis}%[]
%\fnmark[3,4]
\ead{npa4tg@virginia.edu}
\author[5,7,8]{Ingrid Agartz}%[]
%\fnmark[7,8]
\ead{ingrid.agartz@medisin.uio.no}
\author[5,6]{Lynn Egeland Mørch-Johnsen}
%\fnmark[5,6]
\ead{l.e.morch-johnsen@medisin.uio.no}
\author[1]{Graham K. Murray}
%\fnmark[1]
\ead{gm285@cam.ac.uk}
\author[1]{John Suckling}
%\fnmark[1]
\ead{js369@cam.ac.uk}
\author[2]{Pietro Lio}
%\fnmark[2]
\ead{pl219@cam.ac.uk}

% URL of the first author

% Credit authorship
% eg: \credit{Conceptualization of this study, Methodology, Software}
%\credit{}

%\equalcont{These authors contributed equally to this work.}
\affiliation[1]{organization={University of Cambridge, Department of Psychiatry},
            addressline={Hills Road}, 
            city={Cambridge},
%          citysep={}, % Uncomment if no comma needed between city and postcode
            postcode={CB2 2QQ}, 
            state={Cambridgeshire},
            country={United Kingdom}}
\affiliation[2]{organization={University of Cambridge, Department of Computer Science and Technology},
            addressline={15 JJ Thomson Ave}, 
            city={Cambridge},
            postcode={CB3 0FD}, 
            state={Cambridgeshire},
            country={United Kingdom}}
\affiliation[3]{organization={University of Virginia, Department of Environmental Sciences},
            city={Charlottesville},
            state={Virginia},
            country={United States of America}}
\affiliation[4]{organization={University of Virginia, School of Data Science},
            city={Charlottesville},
            state={Virginia},
            country={United States of America}}
\affiliation[5]{organization={(NORMENT), Institute of Clinical Medicine, University of Oslo,},
            city={Oslo},
            country={Norway}}
\affiliation[6]{organization={Østfold Hospital, Department of Psychiatry and Department of Clinical Research},
            city={Grålum},
            country={Norway}}
\affiliation[7]{organization={Centre for Psychiatry Research, Department of Clinical Neuroscience, Karolinska Institutet, Stockholm Health Care Services},
city={Stockholm}, country={Sweden}}
\affiliation[8]{organization={Department of Psychiatric Research, Diakonhjemmet Hospital},
city={Oslo}, country={Norway}}           
% Address/affiliation

% Corresponding author text
\cortext[1]{Corresponding author}

% Footnote text
%\fntext[1]{}

% For a title note without a number/mark
%\nonumnote{}

% Here goes the abstract
%The accelerated progress of artificial intelligence (AI) has popularized deep learning models across domains, yet their inherent opacity poses challenges, notably in critical fields like healthcare, medicine and the geosciences. Explainable AI (XAI) has emerged to shed light on these "black box" models, helping decipher their decision making process. Nevertheless, different XAI methods yield highly different explanations. This inter-method variability increases uncertainty and lowers trust in deep networks' predictions. In this study, for the first time, we propose a novel framework designed to enhance the explainability of deep networks, by maximizing both the accuracy and the comprehensibility of the explanations. Our framework integrates various explanations from established XAI methods and employs a non-linear "explanation optimizer" to construct a unique and optimal explanation. Through experiments on multi-class and binary classification tasks in 2D object and 3D neuroscience imaging, we validate the efficacy of our approach. Our explanation optimizer achieved superior faithfulness scores, averaging 155\% and 63\% higher than the best performing XAI method in the 3D and 2D applications, respectively. Additionally, our approach yielded lower complexity, increasing comprehensibility. Our results suggest that optimal explanations based on specific criteria are derivable and address the issue of inter-method variability in the current XAI literature.

\begin{abstract}
The accelerated progress of artificial intelligence (AI) has popularized deep learning models across various domains, yet their inherent opacity poses challenges, particularly in critical fields like healthcare, medicine, and the geosciences. Explainable AI (XAI) has emerged to shed light on these 'black box' models, aiding in deciphering their decision-making processes. However, different XAI methods often produce significantly different explanations, leading to high inter-method variability that increases uncertainty and undermines trust in deep networks' predictions. In this study, we address this challenge by introducing a novel framework designed to enhance the explainability of deep networks through a dual focus on maximizing both accuracy and comprehensibility in the explanations. Our framework integrates outputs from multiple established XAI methods and leverages a non-linear neural network model, termed the 'explanation optimizer,' to construct a unified, optimal explanation. The optimizer uses two primary metrics—faithfulness and complexity—to evaluate the quality of the explanations. Faithfulness measures the accuracy with which the explanation reflects the network's decision-making, while complexity assesses the comprehensibility of the explanation. By balancing these metrics, the optimizer provides explanations that are both accurate and accessible, addressing a central limitation in current XAI methods. Through experiments on multi-class and binary classification tasks in both 2D object and 3D neuroscience imaging, we validate the efficacy of our approach. Our explanation optimizer achieved superior faithfulness scores, averaging 155\% and 63\% higher than the best-performing individual XAI methods in the 3D and 2D applications, respectively, while also reducing complexity to enhance comprehensibility. These results demonstrate that optimal explanations based on specific quality criteria are achievable, offering a solution to the issue of inter-method variability in the current XAI literature and supporting more trustworthy deep network predictions.
\end{abstract}

% Use if graphical abstract is present
%\begin{graphicalabstract}
%\includegraphics{}
%\end{graphicalabstract}

% Research highlights
\begin{highlights}
\item We introduce a new framework that combines various XAI methods and uses a non-linear "explanation optimizer" to enhance the faithfulness and comprehensibility of deep network explanations.
\item We validate our framework in multi-class and binary classification tasks in 2D object and 3D neuroscience imaging, aiming to address inter-method variability and boost trust in AI predictions.
\item Our approach achieves faithfulness scores 63\% and 155\% higher than the best XAI method in the 2D and 3D applications, respectively, while at the same time reducing complexity and increasing comprehensibility.
\item We propose a generalized version of the framework for any explanation task in a multi-dimensional space. 
\end{highlights}

% Keywords
% Each keyword is seperated by \sep
\begin{keywords}
XAI \sep  neuroscience \sep  brain \sep  3D \sep  2D \sep  computer vision \sep  classification
\end{keywords}

\maketitle

% Main text
\section{Introduction}\label{sec1}
In the last decade, deep learning models have led to transformative changes in various domains, including geosciences \citep{g1,g2}, automotive engineering, healthcare \citep{guide,xai}, and biology \citep{n1,mine2}. However, the inherent opacity of these sophisticated models, often likened to "black boxes," presents formidable challenges in understanding their decision-making processes. In response to this challenge, the field of eXplainable Artificial Intelligence (XAI) has emerged as a pivotal area of research dedicated to explaining these black-box models and elucidating the rationale behind their predictions. Over the years, a multitude of XAI methods have been developed, all aimed at shedding light on how artificial intelligence (AI) arrives at specific predictions.
\par Among others, XAI consists of methods that try to explain complex, nonlinear models by approximating them locally with simple interpretable models, such as local interpretable model-agnostic explanations (LIME ; \cite{lime2}) and SHapley Additive exPlanations (SHAP ; \cite{shap}). In medical imaging, specifically, XAI methods often include attribution techniques like SHAP and Layer-wise Relevance Propagation (LRP ; \cite{lrp}) or sensitivity techniques, with GRAD-CAM being a prominent example \citep{xaimi}. Despite their potential to enhance model trust and learning, XAI methods face significant challenges. For instance, LRP produces smooth results but less accurate explanations due to its focus on positive preactivations, and SHAP struggles with computational intensity in calculating Shapley values (\cite{shap}). Although adaptations such as Monte Carlo methods and stratified sampling (e.g., SVARM) have improved estimation precision \citep{svarm}, it has been established that no single XAI method consistently excels across all prediction tasks \citep{a1}, leading to high variability and uncertainty in deep learning explanations \citep{quan}.
\par
\begin{figure}[!ht]
\centering
\centerline{
\includegraphics[trim={2.cm 0.5cm 2.35cm 0.00cm},clip,width=1.\textwidth]{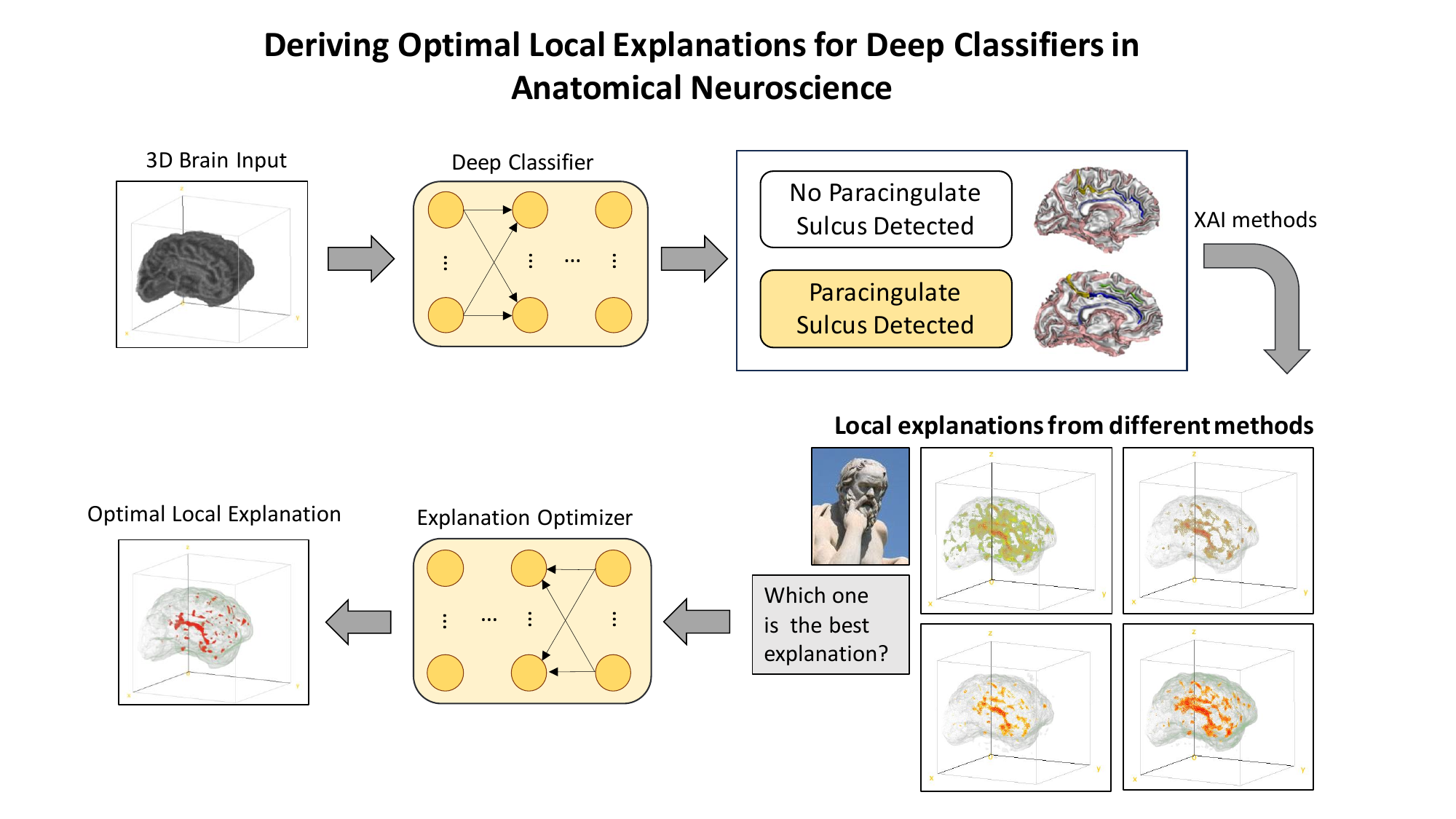}
}
\small
\par\medskip
\caption{Deriving optimal local explanations for deep classifiers in anatomical neuroscience.
The figure depicts the traditional local explanation process (implementing multiple XAI methods) for a three-dimensional binary classification example regarding the existence or absence of PCS (ParaCingulate Sulcus). This is one of the two case studies we consider in this paper. Note that the high variability in the local explanations from different XAI methods increases uncertainty and lowers trust in explainability among experts. Therefore, an explanation optimizer is proposed to derive a unique and optimal explanation for the pre-trained Deep Classifier in the computer vision task of interest.}
\label{fig0}
\end{figure}

 In this study, for the first time, an "explanation optimizer" is proposed to derive an optimal explanation for deep classifiers in computer vision tasks, based on specific pre-defined performance metrics (see Fig. \ref{fig0}), thus, solving the enigma of explainability. The optimizer, a non-linear neural network model, reconstructs a unified explanation by integrating outputs from various XAI methods. This model is guided by two critical metrics: faithfulness, which measures how accurately the explanation reflects the network's decision-making process, and complexity, which assesses how comprehensible the explanation is. By balancing these scores, the optimizer aims to produce explanations that are both true to the model’s internal logic and accessible in their clarity. The framework ultimately provides a unique optimal explanation with the same or higher resolution as the network's input (see Fig. \ref{fig1hh}). We note that providing high resolution explanations is useful for more accurate identification of significant features, particularly crucial in medical applications \citep{surv1}.
To assess the effectiveness of our framework, we conducted experiments on both multi-class and binary classification tasks in two- and three-dimensional spaces, respectively, focusing on automobiles-animals and neuroscience imaging domains. 

The proposed explanation optimizer is inspired by Herbert Simon’s concepts of "bounded rationality" and "satisficing" decision-making \citep{sim,sim2}. "Bounded rationality" reflects the limitations of human knowledge, time, and computational capacity, framing rationality as constrained by heuristics and cognitive biases that aid in efficient yet simplified decision-making. Simon's notion of "satisficing," introduced in 1956, emphasizes finding solutions that are "good enough" rather than fully optimized \citep{sim3}. In our framework, we apply these principles by integrating complementary insights from various XAI methods to achieve explanations that balance high faithfulness with manageable complexity. Our novel cost function supports the idea of "satisficing" explanations, targeting optimal interpretability (as quantified by specific metrics) for specific computer vision tasks across multiple domains, including automobiles, animals, and neuroscience. Furthermore, we leverage the concept of 'bounded rationality' by integrating established knowledge from different XAI methods as inputs to the optimizer, recognizing that these methods may or may not be biased or bounded in different ways.

\par To encapsulate, in this study, we propose a framework aimed at providing optimal explanations for pre-trained deep networks in specific computer vision tasks. 
Characterized by high faithfulness and low complexity, this framework ensures accuracy and comprehension in the provided explanations. Additionally, we develop a separate framework designed to deliver explanations in high-resolution spaces, enhancing our ability to capture and understand significant feature contributions effectively. Our framework is tested in applications across diverse domains, including automobiles, animals, and neuroscience, encompassing both 2D and 3D multi-classification and binary classification tasks. Finally, we formulate a generalized framework for optimizing explanations for M-dimensional inputs, extending the utility of our approach beyond specific computer vision tasks to offer a versatile solution for interpreting complex data across various dimensions and domains. Our results suggest that optimal explanations based on specific criteria are indeed derivable and address the issue of inter-method variability in the current XAI literature.
\section{Related Work}
Explainable artificial intelligence can be broadly classified into two methodological approaches: inherently interpretable and post-hoc. Interpretable methods prioritize models endowed with properties such as simulatability, decomposability, and transparency, making them inherently comprehensible. Typically, these include linear techniques like Bayesian classifiers, support vector machines, decision trees, and K-nearest neighbor algorithms \citep{xai22}. In contrast, post-hoc methods are employed alongside AI techniques in a post-prediction setting to explain the (otherwise non interpretable) AI predictions and unveil nonlinear mappings within complex datasets. In this manuscript, we will focus on post-hoc "local" methods that provide an explanation for each AI prediction separately, as opposed to post-hoc "global" methods that derive explanations for the decision-making of the AI for the entire dataset \citep{surv1}. One notable post-hoc local technique is the local interpretable model-agnostic explanations (LIME), which explains the network's predictions by building simple interpretable models that approximate the deep network locally, i.e., in the close neighborhood of the prediction of interest \citep{lime2}. Post-hoc techniques encompass model-specific approaches designed to address specific nonlinear model behaviors and model-agnostic approaches to investigate data complexity \citep{xai,xai22}. In computer vision, model-agnostic techniques such as LIME and perturbation-based methods find extensive application, while model-specific methods include feature relevance, condition-based explanation, and rule-based learning \citep{lime2,xaims,xai22}. Similarly, in medical imaging, XAI methods predominantly consists of attribution and perturbation techniques \citep{xai_surv}. Attribution techniques identify the important features for a given prediction by assigning relevance scores to features of a given input. Perturbation techniques assess the sensitivity of an AI prediction to certain input features \citep{a1}, by systematically perturbing sub-groups of the input data \citep{xai_surv,xaimi}. Among the XAI methods used in medical imaging, GRAD-CAM emerges as one of the most frequently employed techniques \citep{xaimi}.
\par Despite the benefits and potential of XAI methods in advancing model trust, model fine-tuning and learning new science \citep{a2}, recent studies have shown that they suffer from important pitfalls \citep{a1,a3}. For instance, when using the $\alpha1\beta0$ rule of the method Layer-wise Relevance Propagation (LRP, an attribution method; \cite{lrp}), explanations have been shown to be smooth and comprehensible but of lower accuracy/faithfulness to the deep network decision making \citep{a1}. Reportedly, this is because the $\alpha1\beta0$ rule considers only the positive preactivations in the network \citep{a1,a3}. Recent studies have suggested LRP modifications that combine various LRP rules to produce both faithful and smooth results \citep{lrp2}. As a second example, SHapley Additive exPlanations (SHAP;\cite{shap}), an attribution method based on game theory principles, computes relevance scores called Shapley values for input features, for a given prediction. The primary challenge of implementing SHAP lies in accurately calculating the Shapley values by sampling all theoretically possible combinations of features (i.e., in line with game theory principles), due to constraints raised by computing resources. Several studies have utilized adapted Monte Carlo methods within the permutation space to achieve more precise sampling of Shapley values \citep{mc}. Moreover, efforts have been made to improve results by minimizing the weighted least square error of Shapley values estimation, leveraging learned parametric functions, as described in studies of Kernel-Shap and Fast-Shap \citep{fast}. Similarly, SVARM \citep{svarm} employs stratified sampling by segmenting a dataset into sub-groups or strata and sub-sampling from each input instance. By combining mean estimates from all strata, this technique achieves a more precise estimation of the Shapley values and effectively mitigates the variance. Despite these efforts to optimize already established XAI methods, no existing XAI method has been proven to be consistently superior when considering various performance scores and properties that an ideal method should exhibit (e.g., faithfulness to the network decision making process, comprehensibility, robustness, localization; \cite{quan}). As such, a more holistic approach (implementing many methods) is typically preferred \citep{a1}. However, even when explaining the same deep network and the same prediction, different XAI methods yield highly different explanations (see Fig. \ref{fig0}), increasing uncertainty about the decision making process of deep networks and lowering trust. The sub-optimal results of individual XAI methods coupled with the high inter-method variability constitute one of the biggest challenges in the field of XAI currently and is the focus of this work. 
\section{Methodology}\label{sec11}
\subsection{Evaluation metrics for the explanation}
A crucial aspect of this study lies in evaluating "how accurate and comprehensive is an explanation?" To derive a useful explanation, two primary scores play a pivotal role: faithfulness and complexity. An intuitive way to assess the quality of an explanation is by measuring its ability to accurately capture how the predictive model reacts to random perturbations \citep{fid}. For a deep neural network $f$, and input features $\boldsymbol{x}$, the feature importance scores (also known as "attribution scores") are derived in a way such that when we set particular input features $\boldsymbol{x_s}$ to a baseline value $\boldsymbol{x_s^f}$, the change in the network's output should be proportional to the sum of attribution scores of the perturbed features $\boldsymbol{x_s}$. We quantify this by using the Pearson correlation between the sum of the attributions of $\boldsymbol{x_s}$ and the difference in the output when setting those features to a reference baseline \citep{f}. Thus, we define the faithfulness of an explanation method $g$ as:
\begin{equation}
M_{faith}(f,g;\boldsymbol{x}) = corr_S(\sum_{i\in S} g(f,\boldsymbol{x})_i,f(\boldsymbol{x})-f(\boldsymbol{x}[\boldsymbol{x_s}=\boldsymbol{x_s^f}]))
\label{1}
\end{equation}
where $S$ is a subset of indices ($S\subseteq [1,2,3 ... d]$), $\boldsymbol{x_s}$ is a sub-vector of an input $\boldsymbol{x}$ ($\boldsymbol{x}=\boldsymbol{x_s}\cup \boldsymbol{x_f}$ and $\boldsymbol{x_f}$ the unchanged features of $\boldsymbol{x}$ image). The total number of features, which partition an image is $d$. We denote as $\boldsymbol{x_s}$ the changed features, and $\boldsymbol{x_f}$ the unchanged features of $\boldsymbol{x}$ image. 
\par If for an image $\boldsymbol{x}$, explanation $g$ highlights all $d$ features, then it may be less comprehensible and more complex than needed (especially for large $d$). It is important to compute the level of complexity, as an efficient explanation has to be maximally comprehensible \citep{f}. 
If $P_g$ is a valid probability distribution and the $P_g(i)$ is the fractional contribution of feature $x_i$ to the total magnitude of the attribution, then we define the complexity of the explanation $g$ for the network $f$ as:
\begin{equation}
M_{compx}(f,g;\boldsymbol{x}) = \sum_{i=1}^{d}P_g(i)(log(\frac{1}{P_g(i)}))
\label{4}
\end{equation}
where:
\begin{equation}
P_g(i) = \frac {\vert g(f,\boldsymbol{x})_{i}\vert}{\sum_{j=1}^{d}\vert g(f,\boldsymbol{x})_{j}\vert}
\label{5}
\end{equation}
In order to evaluate these two explainability metrics, we used the software developed by \citep{quan}. This software package is a comprehensive toolkit that collects, organizes, and evaluates a wide range of performance metrics, proposed for XAI methods. We note that we used a zero baseline ('black'; $\boldsymbol{x_s^f}$= $\boldsymbol{0}$), and 70 random perturbations to calculate the Faithfulness score.
\subsection{Overview of the proposed framework}
\par 
The proposed framework entails an optimization approach aimed at attaining the highest faithfulness and lowest complexity of an explanation tailored to a specific deep learning network and computer vision task. 
The framework was applied in 3D and 2D experiments, and the architectures of the optimizers are presented in Fig. \ref{fig1hh} and in Supplementary material Fig. 1, respectively. In both cases, the optimizer is initiated with a layer of local explanations from various already-established XAI methods. A weighted average of these explanations is computed ('Weighted Average'). The individual local explanations and the weighted average are then concatenated and fed into a deep learning network ('Non-linear structure') to reconstruct a unique, optimal explanation. After the optimal explanation is derived, a high-resolution version is extracted by using an up-sampling ('Up-Sampling') layer (see Fig. \ref{fig1hh}). Finally, for training, the cost function evaluates i) the similarity between the weighted average and low-resolution prediction, ii) the faithfulness score of the explanation as quantified in \citep{quan}, and iii) the complexity score of the explanation as quantified in \citep{quan} (Fig. \ref{fig1hh}). The similarity score above (i) is used in the loss function to constrain the optimization solution and to accelerate the training convergence.

\begin{figure}[!ht]

\centering
\centerline{
%\relax \textbf{a}
\includegraphics[width=1.\textwidth]{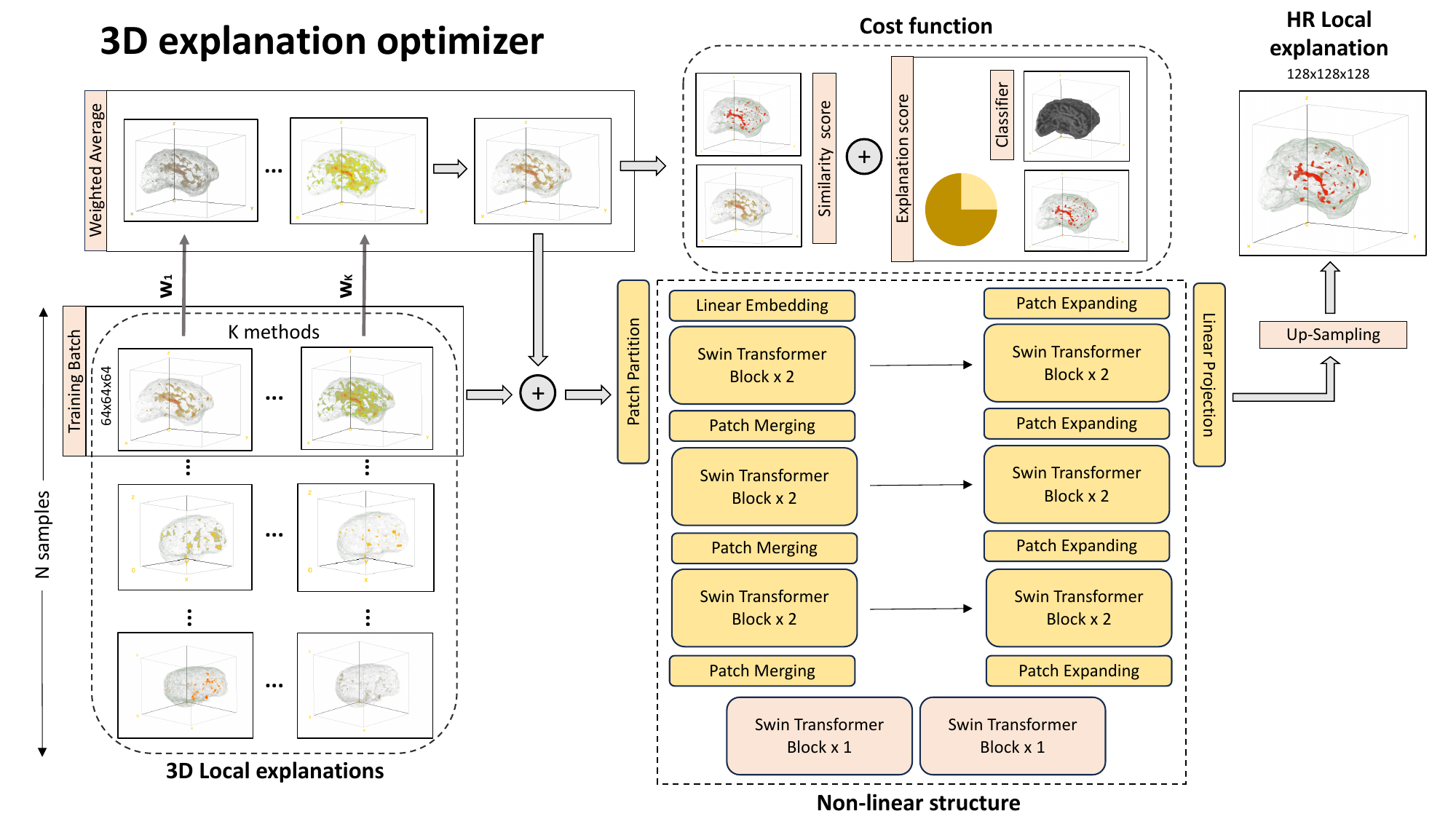}}
%\centerline{
%\relax \textbf{b}
%\includegraphics[width=1.4\textwidth]{pipes2.pdf}}
\small
\caption{The proposed framework for optimizing explanations in the 3D task. The key components of the framework include the utilization of $K$ baseline XAI methods (K methods), a 'Weighted Average' explanation, a non-linear network ('Non-linear structure'), and an 'Up-Sampling' unit. The framework incorporates a range of explanations obtained from established XAI methods and computes an adjusted baseline explanation ('Weighted Average'). The local explanations and the 'Weighted Average' are concatenated and fed into a suitably designed non-linear structure (Swin-UNETR). %(a) and U-Net (b)).
The reconstructed optimal explanation feeds an upsampling deconvolution layer to increase the resolution (high-resolution (HR) local explanation). The cost function evaluates the similarity between the weighted average and the low-resolution prediction, which is added to the faithfulness and complexity scores of the explanation. %\textbf{{a)}}
The 3D application uses images from the Top-Oslo dataset \citep{mo1}.}
%\textbf{{b)}} The 2D application uses images from the CIFAR-10 dataset (\cite{cifar}).
\label{fig1hh}
\end{figure}
\par Key components of the framework include the utilization of the $K$ baseline XAI methods, the 'Weighted Average' explanation, the 'Non-linear structure', and the 'Up-Sampling' unit (Fig. \ref{fig1hh}). The assigned weights of the 'Weighted Average' explanation are computed from the equation below:
\begin{equation}
w_k=l_{1} M_{faith}^k(f,g;\boldsymbol{x}) + l_{2}/M_{compx}^k(f,g;\boldsymbol{x})
\label{eq1}
\end{equation}
where $l_1$ and $l_2$ represent fixed values ranging from 0 to 1, signifying the contributions of faithfulness and complexity, respectively, to the final weighted average. In this study, we analyzed predictions and explanations for ten randomly selected prediction instances from the validation cohort. This selection is made solely for computational efficiency. During training, for each batch, these 10 selected instances serve as a proxy to approximate the relative weights, allowing for the calculation of the weighted average of the explanations. We assessed the faithfulness and complexity scores for each of the $K$ explanation methods and for each prediction instance. Subsequently, we calculated the average faithfulness and complexity scores across the ten prediction instances and determined the final assigned weights using eq. \ref{eq1} with $l_1 = 0.6, l_2 =0.4, $. The average values of $M_{faith}^K(f,g;\boldsymbol{x})$, $M_{compx}^K(f,g;\boldsymbol{x})$ and the assigned weights $w_k$ were normalized to range from 0 to 1. 
\par The $K$ explanations generated by the baseline XAI methods are fed into a non-linear network consisting of an encoder, bottleneck, and decoder architecture (see later sections for more details), with the objective of reconstructing a unique explanation. Following this, the reconstructed explanation undergoes sequential processing through an up-sampling deconvolution layer to double resolution. In our case, the resolutions of the 3D inputs in the  application were $64\times64\times64$, while for the 2D images, they were $32\times32$. The final output of our optimizing framework consists of an up-sampled higher-resolution explanation based on the input ($128\times128\times128$ in the 3D application and $64\times64$ in the 2D application) and an explanation with the same resolution as the input ($64\times64\times64$, and $32\times32$, respectively). The up-sampling approach is motivated by the aim to enhance the detail of the explanation. We note that particularly in neuroscience and medical applications, a higher resolution is closely linked with increased trust of the community to AI procedures, making the provision of explanations in both higher and intial resolutions highly desirable \citep{1,2,3,4}.
\par The down-sampled version of the explanation is evaluated using the faithfulness eq. \ref{1} and complexity score eq. \ref{4}. Both the up-sampled and down-sampled explanations are evaluated as to their similarity to the 'Weighted Average' explanation. These three components (faithfulness, complexity, similarity) together constitute the cost function of the proposed optimization methodology. 
\par The similarity score between the derived explanation and the 'Weighted Average' explanation is as follows:
\begin{equation}
SSIM(\boldsymbol{x},\boldsymbol{y})= \frac{(2 \mu_x \mu_y + c_1)(2 \sigma_{xy}+c_2)}{(\mu_x^2+\mu_y^2+c_1)(\sigma_x^2+\sigma_y^2+c_2)}
\label{1}
\end{equation}
\begin{equation}
loss_{sim}(\boldsymbol{x},\boldsymbol{y})=\lambda_1 (1-SSIM(\boldsymbol{x},\boldsymbol{y})^{LR})+ \lambda_2 (1-SSIM(\boldsymbol{x},\boldsymbol{y})^{HR})
\label{2}
\end{equation}
where $\boldsymbol{x}$ represents the derived explanation by our optimizer, $\boldsymbol{y}$ denotes the 'Weighted Average' explanation, $\mu_{x}$ indicates the average of $\boldsymbol{x}$, $\sigma_x^2$ signifies the variance of $\boldsymbol{x}$, $\sigma_{xy}$ represents the covariance of $\boldsymbol{x}$ and $\boldsymbol{y}$, and $c_1$ and $c_2$ are two parameters utilized to stabilize the division with a weak denominator (\cite{ssim}). $SSIM(\boldsymbol{x},\boldsymbol{y})^{LR}$ refers to the similarity score between the 'Weighted Average' explanation and the down-sampled predicted explanation (low resolution), while $SSIM(\boldsymbol{x},\boldsymbol{y})^{HR}$ signifies the similarity score between a tri-linear or bicubic (in 3D and 2D space respectively) up-sampled 'Weighted Average' explanation and the derived high-resolution explanation. The values of $\lambda_1$ and $\lambda_2$ were fixed at 0.5 each. 
The total loss function was given by:
\begin{equation}
loss_{total}(\boldsymbol{x},\boldsymbol{y})=-l_{1} M_{faith}(f,g;\boldsymbol{x}) + l_{2} M_{compx}(f,g;\boldsymbol{x})+ l_3 loss_{sim}(\boldsymbol{x},\boldsymbol{y}) 
\label{3}
\end{equation}
Following an ablation study exploring various combinations of the values for $l_1, l_2, $ and $l_3$ in the above loss function, we determined that setting $l_1 = 0.5, l_2 =0.3, $ and $l_3 = 0.2$ is optimal. This selection ensures a balanced optimization approach that effectively addresses the unique challenge of deriving an optimal explanation. 
\subsection{The baseline XAI methods}
For this study, we employed eight established XAI methods ($K=8$): Saliency \citep{shap}, DeepLift \citep{shap}, Kernel Shap \citep{shap}, DeepLift Shap \citep{shap}, Integrated Gradients \citep{int}, Guided Backpropagation \citep{grad}, Guided GradCam \citep{cam} and Gradient Shap. GradientShap makes an assumption that the input features are independent and that the explanation model is linear, meaning that the explanations are modeled through the additive composition of feature effects. Under these assumptions, SHAP values \citep{shap} can be approximated as the expectation of gradients computed for randomly generated samples from the input dataset. Gaussian noise is added to each input, creating different baselines for computing the SHAP values. \par Extending the implementation of these explainability approaches from two-dimensional (2D) to three-dimensional (3D) space was necessary. We utilized the Python software developed by \citep{quan} as the underlying libraries for our implementation. Further implementation details can be found in the public GitHub repository associated with this study. It's important to note that all explanations were modified to focus exclusively on positive attributions. Any negative attribution in the original explanations were replaced with a zero value. As part of future work, we plan to incorporate negative attributions into the reconstruction process of the explanation optimizer to enhance interpretability.
\subsection{Deep learning architectures}
We utilized a Resnet-18 and a ResNet-50 network architectures \citep{r1} for the classification tasks in the three-dimensional (3D) and two-dimensional (2D) applications, respectively. 
\par With regards to the explanation optimizer, for the 3D application, we employed the Swin-UNETR architecture \citep{rec} as our non-linear reconstruction network, featuring a 24-feature size (Fig. \ref{fig1hh}). For the 2D explanation optimization, we utilized a U-NET encoder-decoder architecture \citep{u1}; see Fig. 1 in the Supplementary material). 
\par For the sake of further investigation (see subsection 4.5.'Why utilizing a non-linear optimizer structure over linear alternatives?'), we explored a linear implementation in the three-dimensional application. Here, we employed visual transformer encoders \citep{vit} to estimate weights for each of the $K$ explainable baseline methods (refer to Supplementary material subsection: 'Deep learning architectures: The linear implementation of the proposed network'). Subsequently, these estimated weights were utilized in a weighted average to extract the linearly optimized explanation.
\section{Experiments and Results}\label{sec2}
\subsection{The experiments in 3D and 2D applications}
We assess the proposed framework in two distinct spatial domains: three-dimensional (3D) and two-dimensional (2D). Our objective is to evaluate the applicability of the proposed framework across various dimensionalities of computer vision inputs. 
\par In the 3D application, our focus is on a binary classification task concerning the presence or absence of the paracingulate sulcus (PCS) in the brain. Leveraging a comprehensive cohort of 629 subjects from the TOP-OSLO study \citep{mo1}. We utilized data from 596 patients after excluding 33 cases due to pre-processing errors in 3D topology extraction and inadequate PCS detection levels. For the 3D classification task, we utilize white-grey surface matter images extracted from T1-weighted MRI scans of patient brains. Our image preprocessing follows the outlined steps in \citep{our1}.
For the 2D application, we utilize the publicly available CIFAR-10 dataset (\url{https://paperswithcode.com/dataset/cifar-10}). This dataset, a subset of the Tiny Images dataset, consists of 60,000 colored images with dimensions of $32\times32$. Each image depicts and corresponds to one of 10 mutually exclusive labels (i.e., classes), such as airplane, automobile, bird, cat, deer, dog, frog, horse, ship, and truck.
\subsection{Data statement for the 3D TOP-OSLO dataset}
The data supporting the findings of the present study are hosted in the repository at NORMENT/Oslo University Hospital. Restrictions apply to the availability of data and they are thereby not publicly available. Data can be made available under reasonable request and with permission of NORMENT/Oslo University Hospital, in accordance with the ethics agreements/research participants consent.
\subsection{Implementation details}
For both the 2D and 3D classification tasks, after randomly shuffling the data, each dataset underwent splitting into training, validation, and testing sets, with proportions of $60\%$, $20\%$, and $20\%$ of the total number of images, respectively. Sparse categorical cross-entropy was employed as the cost function, while the Adam algorithm \citep{adam} was utilized to optimize the loss function. Initially, the learning rate remained fixed for the first 300 epochs, followed by a decrease of $0.1$ every 100 epochs. To prevent overfitting, an early stopping criterion of 100 consecutive epochs was enforced, with a maximum of 500 epochs used for the input modalities in the corresponding 2D and 3D applications. Data augmentation techniques were applied, including rotation (around the center of the image by a random angle from the range of $[-15^{\circ}, 15^{\circ}]$), width shift (up to $20$ pixels), height shift (up to $20$ pixels), and Zero phase Component Analysis (ZCA;\cite{noise}) whitening (adding noise to each image). Hyperparameter tuning was conducted for both applications using four different values of initial learning rates: $1e^{-2}$, $1e^{-3}$, $1e^{-4}$, and $1e^{-5}$. 
\par The same fixed-step learning rate and optimization algorithm, namely the Adam algorithm \citep{adam}, were employed to optimize the loss function in the explainability task. However, no data augmentation techniques were utilized to train the optimizer. The cost function used was that of eq. \ref{3}. For the 2D task, the maximum number of epochs was set at 150, with an early stopping criterion of 10 epochs applied after passing the first 90 epochs. For the 3D task, the maximum number of epochs was set at 500, with an early stopping criterion of 120 epochs applied after the initial 300 epochs. Once again hyperparameter tuning was conducted for both applications using four different values of initial learning rates: $5e^{-2}$, $5e^{-3}$, $5e^{-4}$, and $5e^{-5}$. 
\par The code developed in this study is written in the Python programming language using pyTorch (Python) libraries. For the training and testing of deep learning networks, we have used an NVIDIA cluster with $4$ GPUs and $64$~GB RAM memory. We used an NVIDIA A100-SXM-80GB GPU, and the code will be publicly accessible through github. 
\subsection{Results of classification training and optimization of explanations}
The deep neural networks were trained for binary and multi-label classification tasks, in three and two dimensional spaces. The sensitivity of the results for the different learning rates is illustrated with light blue shading in Fig. \ref{fig3ee} and in Supplementary material Fig. 3. Results show the minimum, median and maximum loss and accuracy values that the deep networks achieved in the training and validation datasets for the 3D and 2D applications, respectively. In the 3D binary classification task, the network achieved approximately 75.0\% validation accuracy and around 70.0\% training accuracy (see Fig. \ref{fig3ee}). For the 2D multi-classification task, the highest validation accuracy exceeded 80.0\%, with training accuracy surpassing 75.0\% (see Supplementary material Fig. 3). Fig. \ref{fig3ee} and Supplementary material Fig. 3 demonstrate the effective training of Resnet-18 and ResNet-50 models on Top-OSLO and CIFAR-10 datasets, respectively. The training and validation performance is shown to be fairly similar, indicating a low risk of overfitting.  We note that the optimal learning rate for the three-dimensional application was found to be $1e^{-4}$, while for the two-dimensional application, it was $1e^{-2}$.
\par Next, Fig. \ref{fig31} and Supplementary material Fig. 4 present the loss function during training of the explanation optimizer for each class in the 3D and 2D tasks. The light blue shading in Fig. \ref{fig31}a. and Supplementary material Fig. 4 indicates the min and max loss values, while the solid currve represents the median loss across different learning rates. The best performance for the optimizer was achieved in the three-dimensional application for a learning rate of $5e^{-3}$ for the presence of PCS, $5e^{-4}$ for the absence of PCS, and $5e^{-3}$ for the two-dimensional scenario.
\begin{figure}[!ht]
\centerline{
\relax \textbf{a}
\boxed{\includegraphics[width=0.5\textwidth]{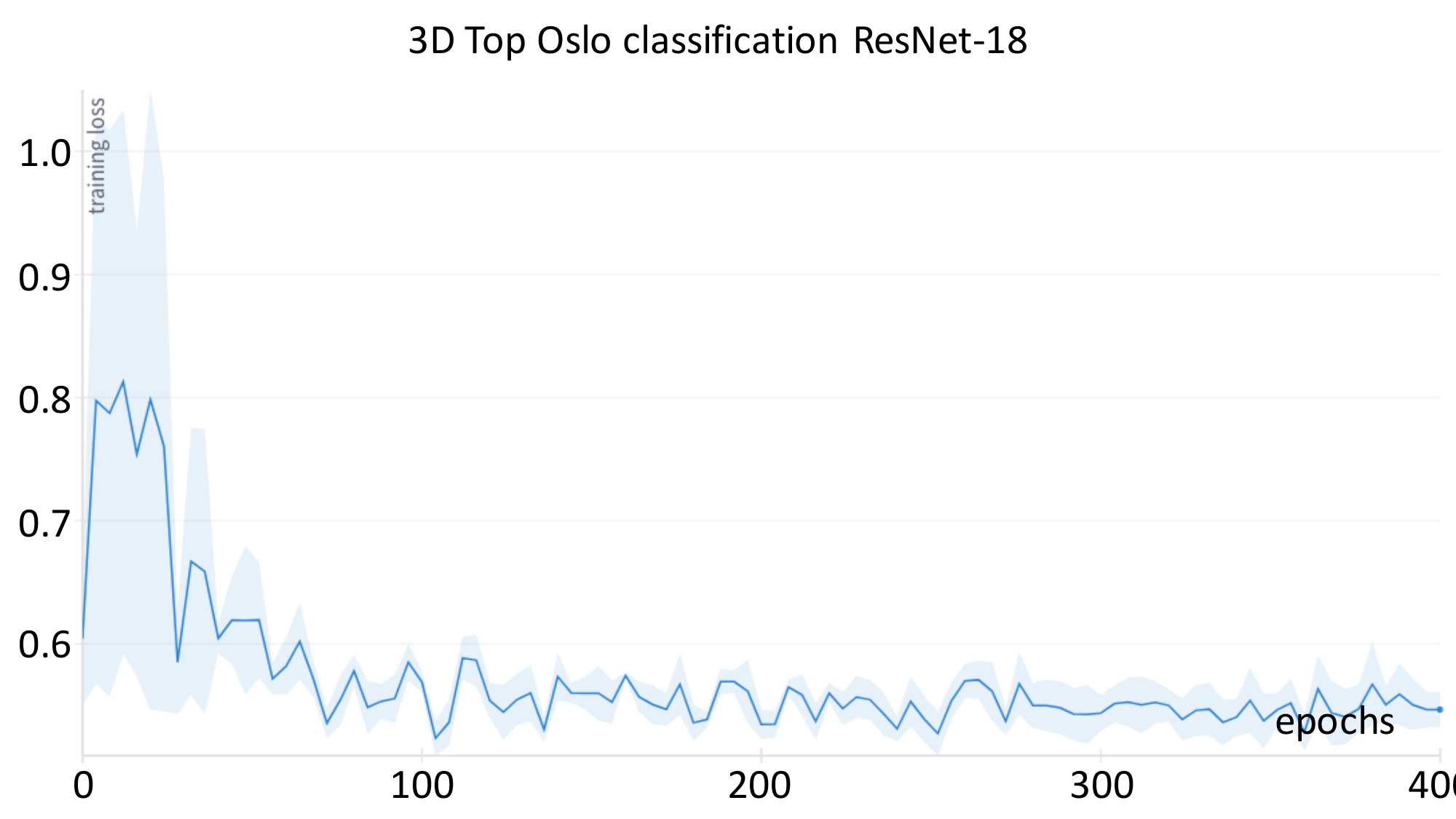}}
\boxed{\includegraphics[width=0.5\textwidth]{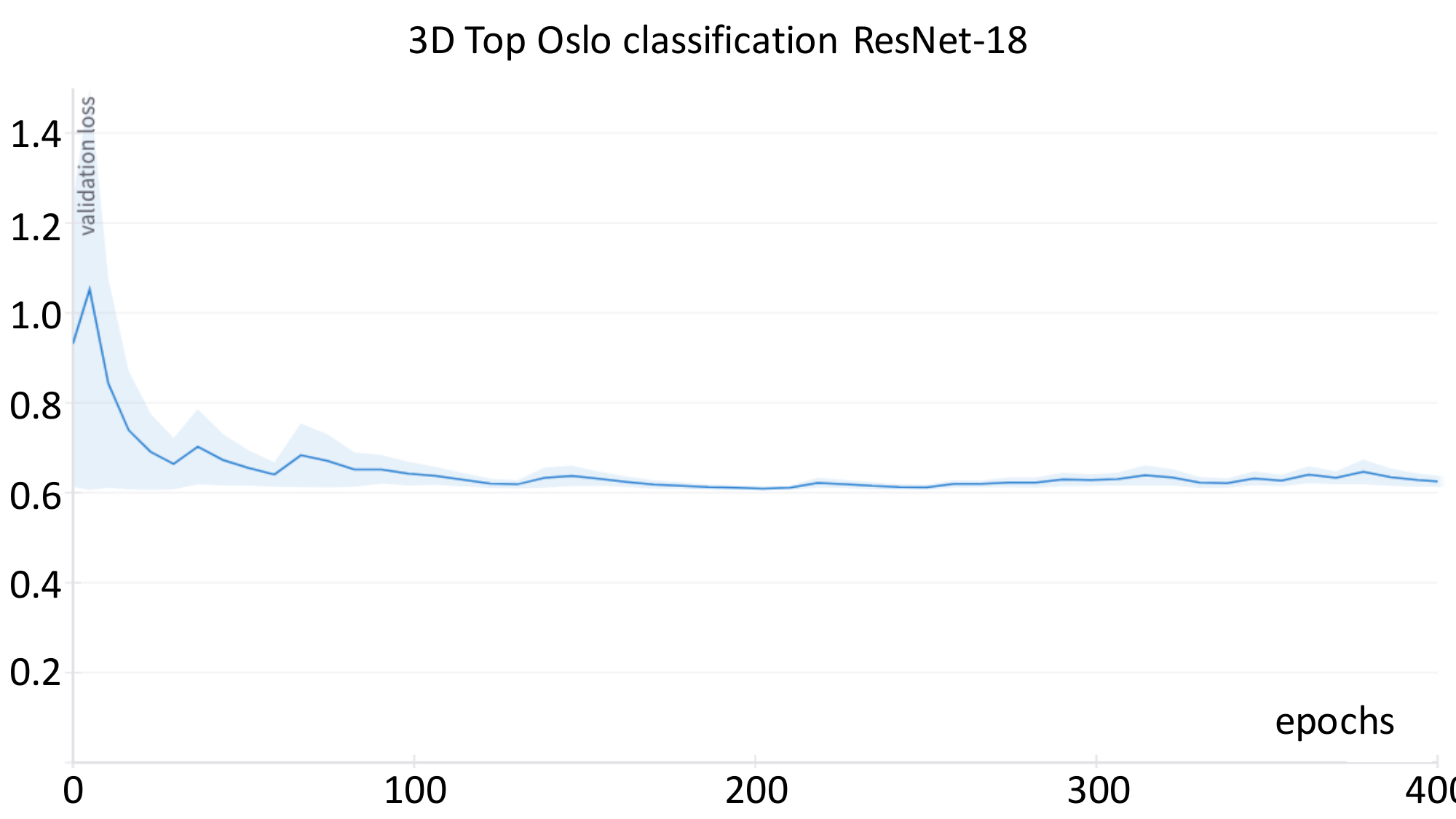}}}
\centerline{
\relax \textbf{b}
\boxed{\includegraphics[width=0.5\textwidth]{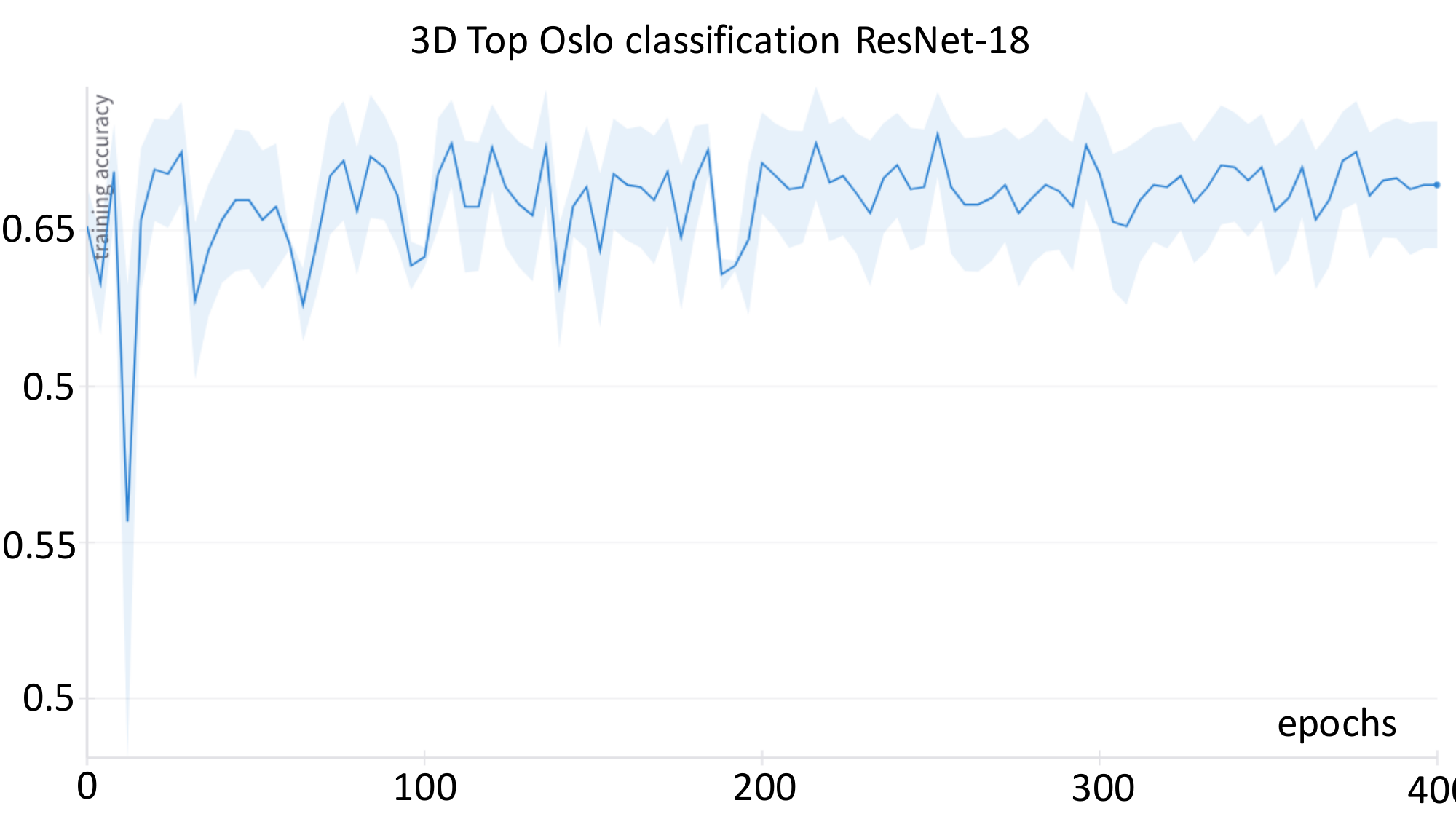}}
\boxed{\includegraphics[width=0.5\textwidth]{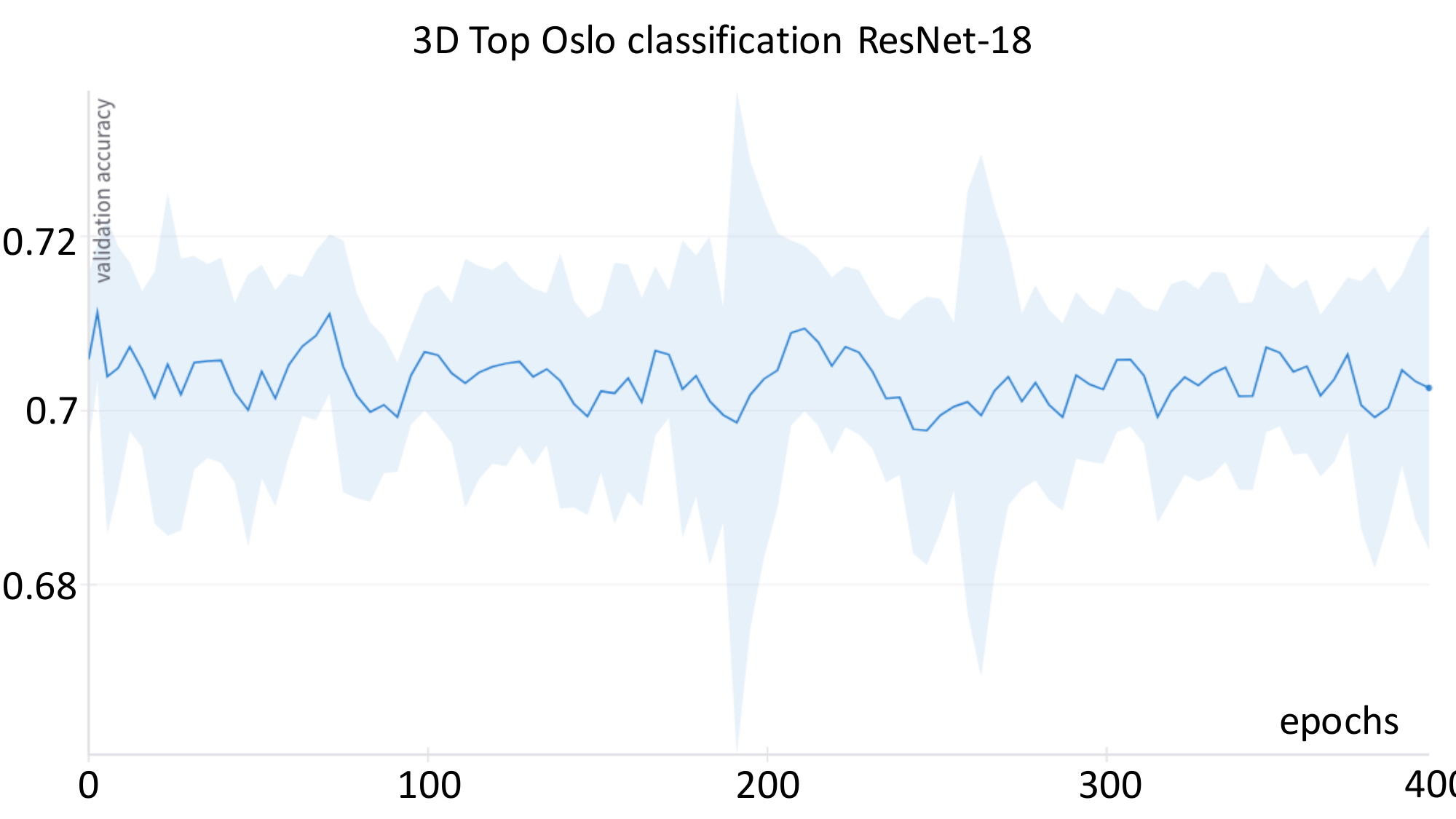}}}
\small
\caption{Training results for the 3D classification task: Detecting the existence/absence of PCS.
\textbf{(a)} The curves and light blue shading in the panels highlight the median and min/max loss values, respectively, obtained during training across various learning rates. The utilized network is Resnet-18, trained and validated in the Top-Oslo cohort. The loss based on training (validation) data is shown on the right (left) panel.  \textbf{(b)} Same as in (a), but the classification accuracy is presented. }
\label{fig3ee}
\end{figure}
\subsection{Explainability scores of the explanation optimizer}
Fig. \ref{fig41} and Supplementary material Fig. 5 present the faithfulness and complexity scores obtained using various XAI methods alongside our proposed non-linear explanation optimizer for the three-dimensional and two-dimensional applications, respectively. In terms of faithfulness, our approach consistently outperforms other XAI methods, achieving higher average scores (exceeding 0.20 in the 3D application and 0.17 in the 2D application). Regarding complexity, our proposed approach demonstrates lower average values (below 12.15 in the 3D application and 9.34 in the 2D application), thus leading to more comprehensible explanations. Regarding the 2D application, we briefly mention that the most faithful XAI method following our approach is DeepLiftShap, achieving an average score of 0.07, while the simplest XAI method behind our optimizer is KernelShap, with an average complexity score of 9.38.
\par We have performed statistical analysis (see Figure 11 and section 5 in Supplementary material) using a two-tailed ANOVA and normality tests across four different scenarios. The ANOVA results show highly significant differences between groups, with F-values of 147, 17.01, 1142, and 37.35, and corresponding p-values all less than 2e-16, indicating that the group differences are statistically significant. However, the Shapiro-Wilk tests indicate significant deviations from normality, with p-values below 2.2e-16 in some cases, suggesting that the normality assumption for ANOVA may not hold. To address this concern, a nonparametric Kruskal-Wallis test was also conducted, yielding p-values of 3.3e-65, 3.22e-18, 9.32e-22, and 3.29e-18, which also support the conclusion of significant differences between groups, despite the non-normality in the data. Based on these results, we can confirm the significant improved performance of our proposed explanation optimizer across all the different experiments.
Overall, these results verify that the proposed optimizer yields more optimal explanations than alternative XAI methods both in terms of faithfulness to the deep classifier and in terms of complexity, improving comprehensibility. Thus, our results provide new evidence that optimal explanations are indeed derivable and point to a new direction in XAI applications - that of combining existing methods to achieve higher level understanding of deep networks. 
\par Fig. \ref{fig5} and Supplementary material Fig. 6 a, present specific explanations in the form of heatmaps, highlighting significant features for both the 3D and 2D applications. In the 3D application, Fig. \ref{fig5} a., b. present two different examples where the Resnet-18 has successfully identified the absence and existence of PCS in the brain, respectively. The different panels show the important features of the 3D brain (sulci sub-regions in the medial view) that helped the netwotk predict, as identified by different XAI methods and our explanation optimizer. In each case, all important features are presented in the first row, and the top 10\% important features are shown in the second row. A significant variability is observed across the explanations provided by existing XAI methods like DeepLift, Kernel Shap, Gradient Shap etc., making it challenging to determine which the optimal explanation is. The explanation from our proposed optimizer ('Explanation Optimizer (non-Linear)') highlights the sub-regions of the areas where the PCS is usually expected to be located \citep{our1,protocol} (Fig. \ref{fig5} a. b). On the contrary, other XAI methods highlight areas that are not as typically related to the PCS connectomics (Gradient Shap; Fig. \ref{fig5} a.,b. and DeepLift Shap; Fig. \ref{fig5} b.). 
\begin{figure}[!ht]
\centerline{
%\relax \textbf{a}
\boxed{\includegraphics[width=0.5\textwidth]{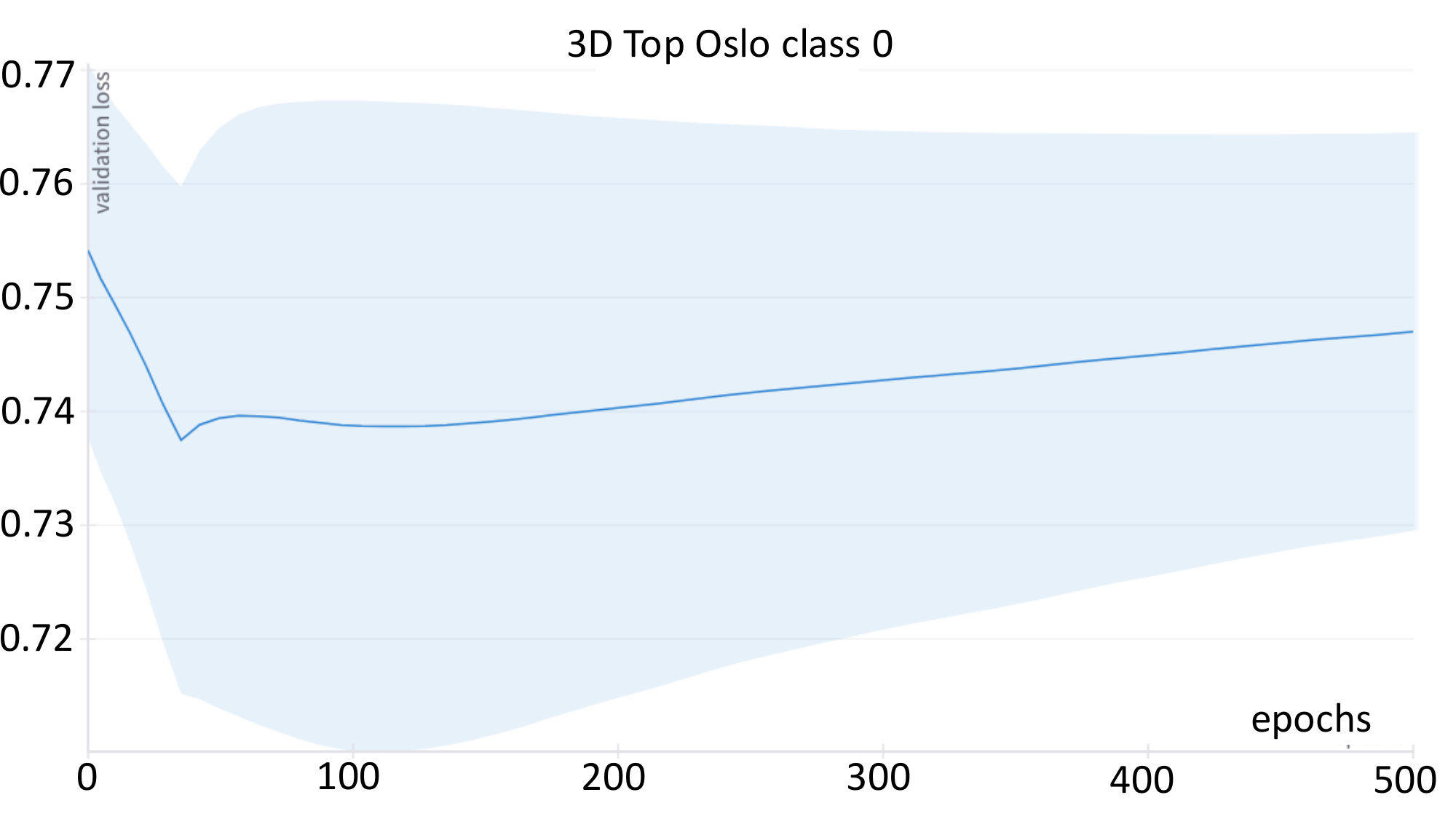}}
\boxed{\includegraphics[width=0.5\textwidth]{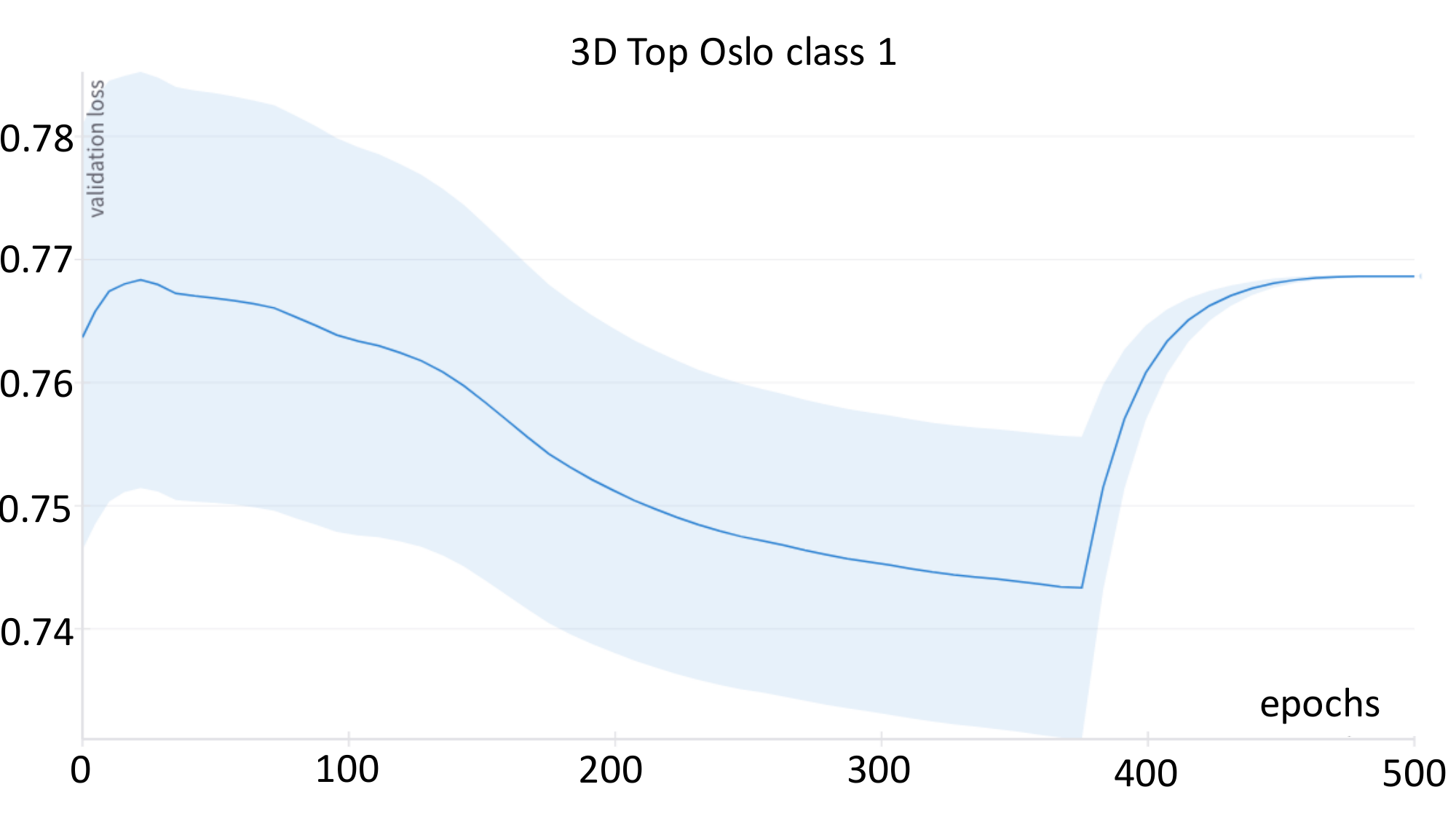}}}
\small
\caption{Explanation optimization for the 3D application:  Training Results.
The curves and light blue shading in the panels highlight the median and the min/max loss values, respectively, obtained during training across various learning rates. The two panels present the validation loss of the explanation optimizer for the binary classification task for class 0 (absence of PCS; left panel) and class 1 (existence of PCS; right panel) in the 3D brain application.}
\label{fig31}
\end{figure} 
\begin{figure}[!ht]
\centering
\centerline{
%\relax \textbf{a}
\includegraphics[trim={0.25cm 3.5cm 0.25cm 2.50cm},clip,width=1.\textwidth]{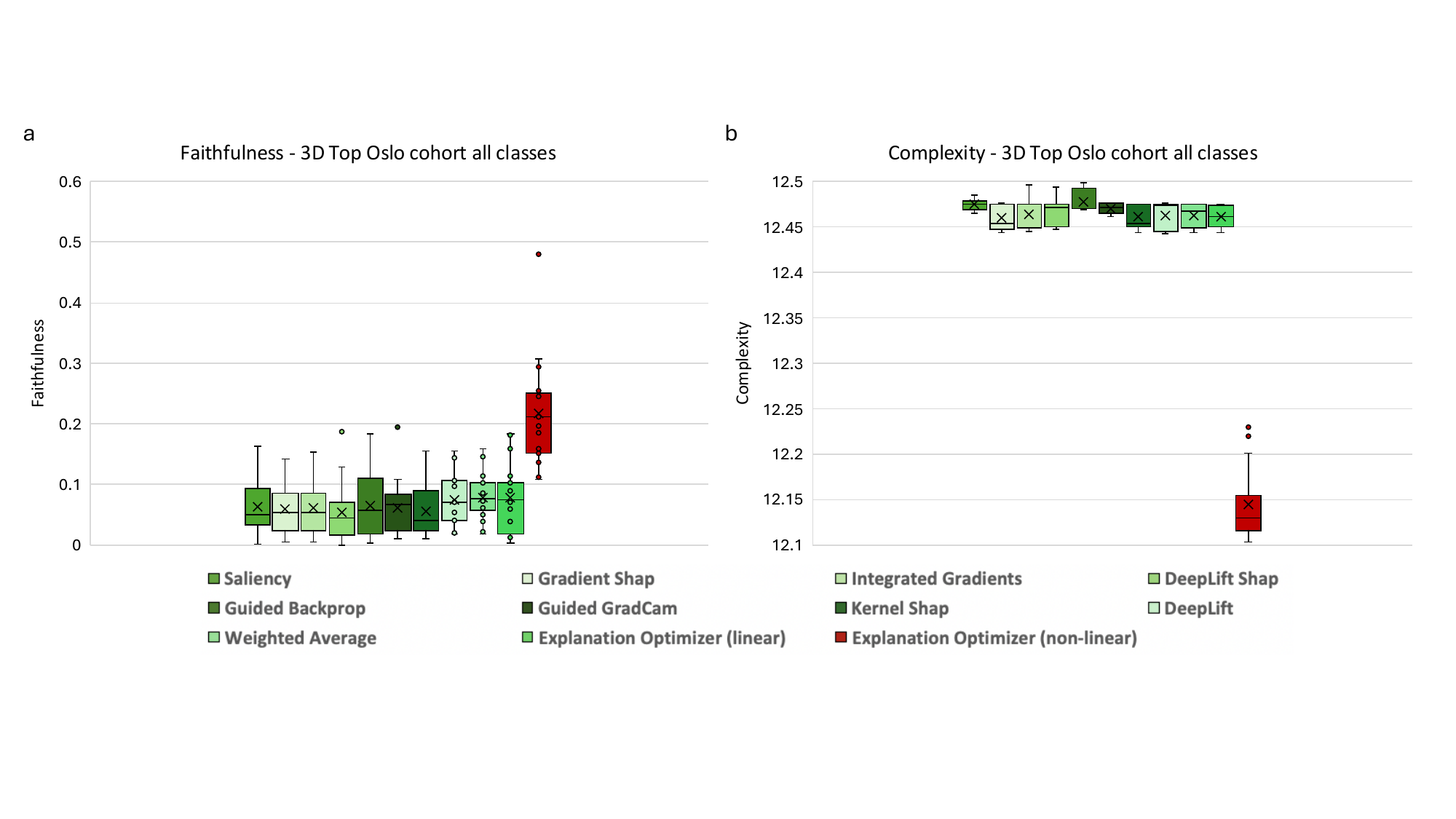}}
%\centerline{
%\relax \textbf{b}
%\includegraphics[width=1.2\textwidth]{box2.pdf}}
\small
\par\medskip
\caption{Box plot results comparing state-of-the-art XAI methods with the proposed explanation optimizer for the 3D application in the testing cohort.
\textbf{(a)} The faithfulness score of different state-of-the-art XAI methods (green color variation) alongside the proposed non-linear explanation optimizer (red color) for the 3D application. \textbf{(b)} Same as in (a), but for the complexity score.}
\label{fig41}
\end{figure}
\begin{figure*}[p]
\centering
\centerline{
\relax \textbf{a}
\includegraphics[width=1.\textwidth]{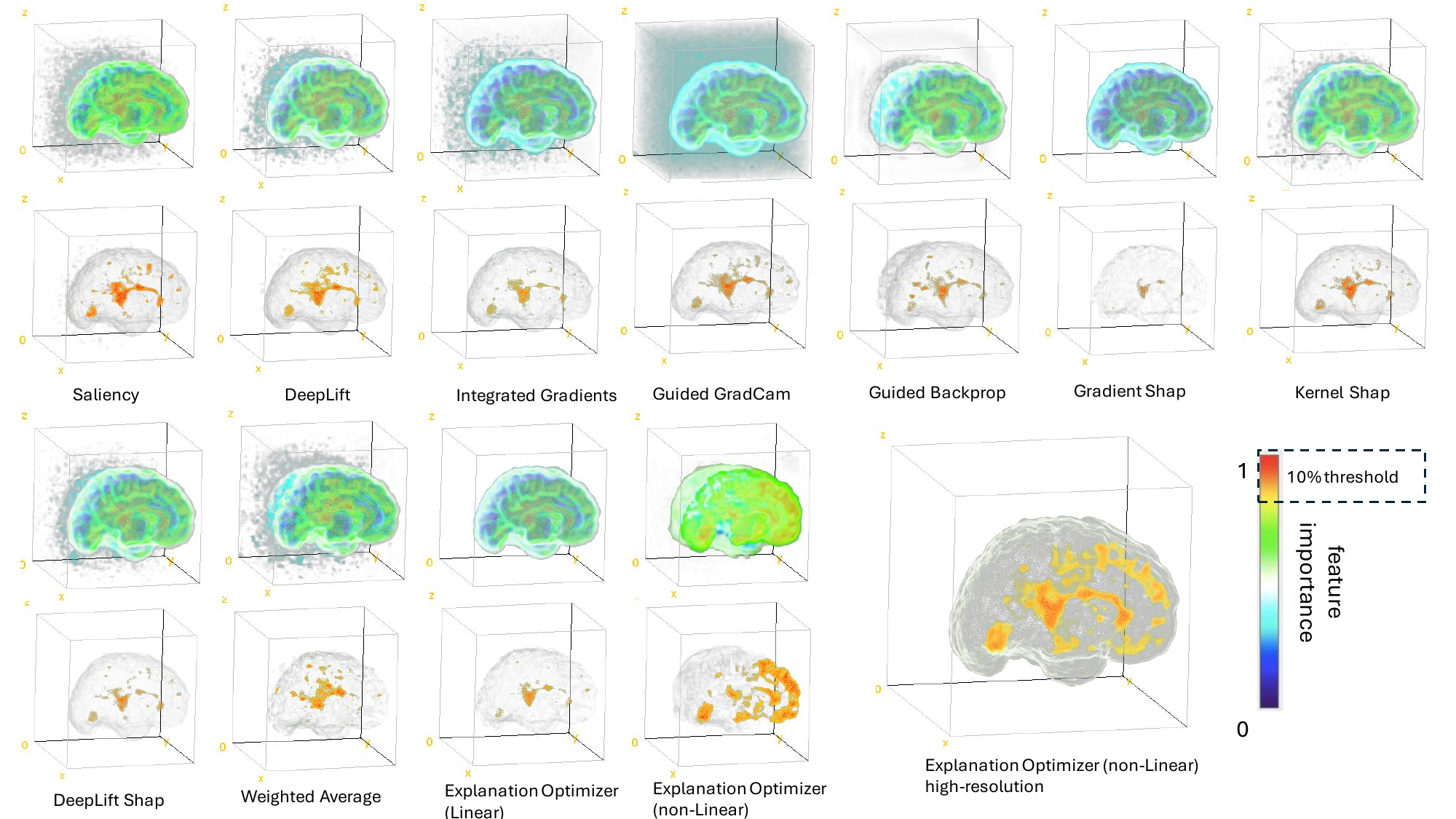}}
\centerline{
\relax \textbf{b}
\includegraphics[width=1.\textwidth]{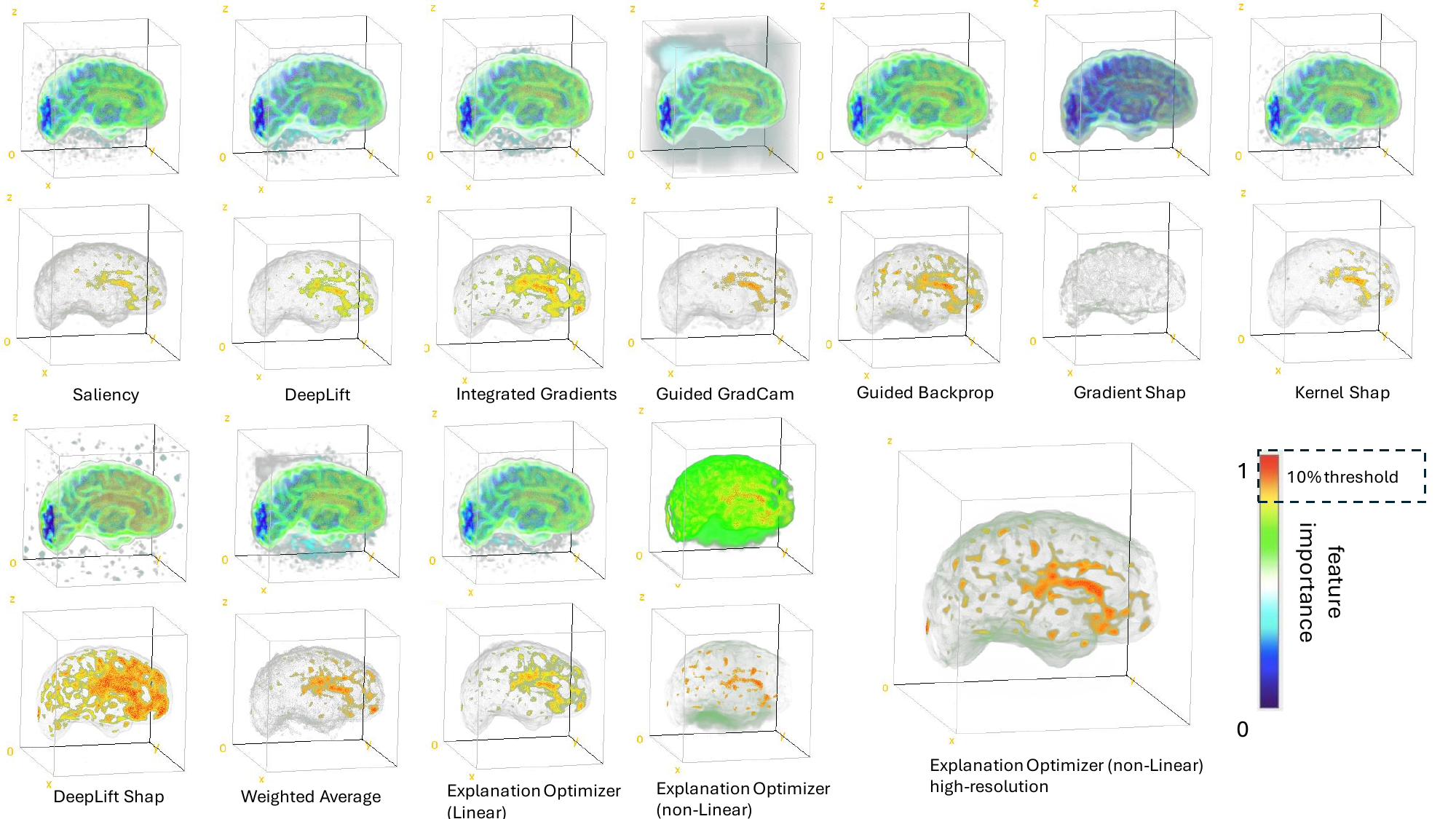}}
\small
\caption{Visualization of local explanations generated based on state-of-the-art XAI methods and the proposed explanation optimizer for the 3D application.
\textbf{(a)} The panels present the local explanations generated from different state-of-the-art XAI methods, alongside the results from the proposed explanation optimizer (the results of both linear and nonlinear implementations are presented), for the 3D application. Explanations are for a specific human individual, randomly selected from the test cohort of class 0 (absence of PCS). For each XAI method, the top row highlights the feature importance using a color scale from 0 to 1 (blue to red), while the second row highlights the most significant features (top 10.0\%). \textbf{(b)} Same  as (a), but results are for a specific human individual, randomly selected from the test cohort of class 1 (existence of PCS).}
\label{fig5}
\end{figure*}
Thus, the proposed explanation optimizer seems to deliver more realistic patterns than other XAI methods regarding which sub-regions contribute to predicting the existence or absence of PCS \citep{our1}. Also, the explanation from our optimizer exhibited the highest scores of faithfulness (see also Fig. \ref{fig41}.a), identifying regions of significance that align with results from the study \citep{our1}.
Last, the proposed approach demonstrated less complexity (see also Fig. \ref{fig41}.b) and focused on specific sub-regions (Fig. \ref{fig5} a., b.), advancing the understanding of the underlying mechanisms.
\par For the two-dimensional examples, explanations of many different 
prediction instances (airplanes, cats, birds, cars, dogs, frogs, and deers) are presented. For simplicity, only the top 10\% important features are shown; see Supplementary material Fig. 6 a. The results suggest that our proposed method provides more accurate explanations of the network's predictions compared to other XAI methods, emphasizing objects and foreground parts that are expected to be critical for classification. For instance, in the first airplane example (top row), saliency and DeepLift primarily focus on the airplane's head, Kernel Shap focuses on the tail, and Gradient Shap focuses on the body and head, while our proposed optimizer identifies all parts of the air plane as significant features. Similar remarks are valid for the car classification, where the optimal explanation very explicitly identifies all the body parts of the car as important features to the network prediction. 
\subsection{Why utilizing a non-linear optimizer structure over linear alternatives?}
A question that may arise is why we opted to use a non-linear structure in our proposed framework (see Fig. \ref{fig1hh} and Fig. \ref{fig2g}) to reconstruct the unique and optimal explanation for the deep learning network, instead of using a linear structure. To investigate this further, we explored a linearly structured optimizer (see Supplementary material Fig. 2) and compared the results to those of the non-linear one. Specifically, in the 3D neuroscience application, where the complexity of the space is higher, we decided to further examine reconstruction structures beyond the non-linear approach. Therefore, we employed a linear structure with attention mechanisms to estimate scalable weights to be used in a weighted average of the existing XAI methodologies. The baseline 'weighted average' mentioned in the Methods section and presented in Fig. \ref{fig2g} is considered as another linear approach employing fixed weights. Both linear approaches, whether employing fixed weights ('Weighted Average') or trainable weights ('Explanation Optimizer (Linear)'), yielded comparable results to the existing XAI methods, as indicated in Fig. \ref{fig41} a. and b. This suggests that linear solutions may not significantly outperform existing approaches in optimizing explanations for deep learning networks. Consequently, there is a rationale for exploring non-linear approaches, as done in this study.

A sensitivity analysis comparing the linear and non-linear structures for different learning rates is provided in the Supplementary material, entitled 'Sensitivity analysis for linear and non-linear structures' (see Supplementary Fig. 8). The main result is that the linear structure is more constrained in faithfulness and complexity scores during training, ultimately converging to lower faithfulness and higher complexity scores compared to the non-linear structure. In contrast, the nonlinear structure exhibits a more stochastic behaviour during training, ultimately converging to a significantly more optimal solution.
\subsection{Justifying the pursuit of optimal explanations for deep learning networks}
In this study, we propose a method to address the challenge of identifying the optimal explanation among various XAI methods. We acknowledge in the manuscript that while adopting a holistic approach and using multiple XAI methods has been shown to be beneficial in some settings \citep{a1,a3,a2}, it can result in highly different explanations, sometimes yielding overwhelming and incomprehensible outcomes, thus, lowering the trust to the network's predictions. Our study demonstrates that opting to derive optimal explanations based on user-defined explainability quality metrics is feasible (see Fig. \ref{fig41} and Supplementary material Fig. 5) and a promising alternative. Visual examples of explanations validate this claim, reinforcing the credibility of such an approach (see Fig. \ref{fig5} and Supplementary material Fig. 6).
\par Deriving optimal explanations has been successfully achieved here by utilizing an innovative loss function. As a first component of the loss, we used the two scores of faithfulness and complexity, so as to ensure that the derived explanation is faithful to the network's prediction-making process and maximally comprehensible. Additionally, we used a third term in the loss representing the similarity between the derived explanation and a baseline weighted average of explanations from existing XAI methods. By employing this similarity term, we further constrained the optimizer's training, promoted convergence, and mitigated divergence toward outlier or trivial solutions (e.g., red noise, impulse noise). By adjusting the importance of each one of these loss terms, one may formulate a unique cost function tailored to solve the problem of optimal explanation based on the needs of their application. We note that in our study, we chose the relative importance of these loss terms via hyper-parameter tuning (see Methods).
\par
\begin{figure*}[!ht]
\centering
\centerline{
\includegraphics[trim={0.25cm 0.5cm 0.0cm 0.00cm},clip,width=1.\textwidth]{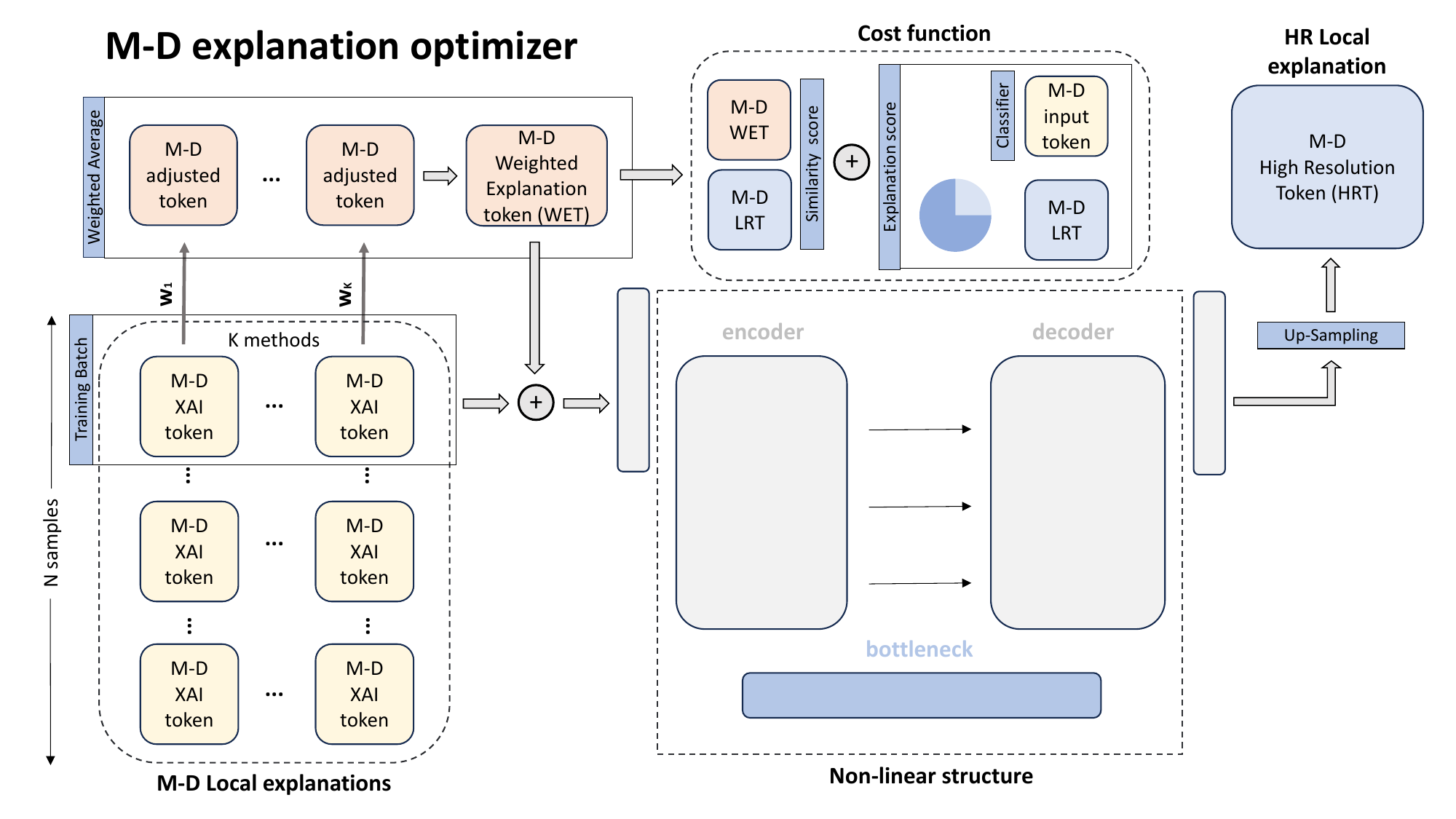}}
\small
\caption{The proposed framework for deriving optimal explanations in a M-dimensional space. The generalized version of the proposed explanation optimizer in M dimensions. \textbf{M:} the dimensionality of the input space, \textbf{K:} the number of XAI methods used as baseline, \textbf{input token:} the input features for the AI task, \textbf{XAI token:} The local explanations from the baseline methods, \textbf{$w_1..w_K$:} The assigned weights based on the scoring of each XAI method on different explainability metrics (here complexity and faithfulness), \textbf{adjusted token:} The product of $w_1,w_2,...w_K$ and the XAI token, \textbf{Weighted Explanation Token (WET):} the weighted average based on the K adjusted tokens of XAI methods, \textbf{High Resolution Token (HRT)} represents the high resolution version of the optimal explanation after the 'non-linear structure' layer. \textbf{Low Resolution Token (LRT)} represents the low resolution optimal explanation, that is, it refers to the resolution of the input token. }
\label{fig2g}
\end{figure*}

 The final component of the proposed framework involved an Up-Sampling layer, aiming to provide a higher resolution explanation than the resolution of the prediction task (Fig. \ref{fig2g}). This higher resolution explanation is vital, particularly in the domain of neuroscience, as it allows for the capture of more intricate details, enhancing the accuracy and comprehensiveness of the explanation. Fig. \ref{fig5}a. and b. show that the results of the 'Explanation Optimizer (non-Linear) high-resolution' are slightly different from the results of the 'Explanation Optimizer (non-Linear)'. The high resolution explanations are more in alignment with the given protocol for the identification of PCS (\cite{protocol,our1}), supporting that the Up-sampling layer is a beneficial addition to our approach.
\subsection{Beyond the two and three dimensions}
We demonstrate our proposed framework in both 3D and 2D applications (Fig. \ref{fig1hh} and Supplementary material Fig. 1, respectively). For the sake of generalization of our proposed framework, we extended it to \textit{'M'} dimensions (i.e., referring to the input space), based on the principles outlined in our approach (see Fig. \ref{fig2g}). This generalized version can be adapted to any Artificial Intelligence application. Specifically, we denote \textit{'K'} different XAI methods used as a baseline, with the \textit{'input token'} representing the input features in a specific AI task of interest (see Fig. \ref{fig2g}).
\par The \textit{'XAI token'} represents the essential features needed for the local explanation, generated by the baseline XAI methods. Furthermore, we denote different assigned weights \textit{$w_1 ... w_k$} derived from explainability metrics such as complexity and faithfulness, and we use the term \textit{'adjusted token'} to refer to the products of the weights $w_1, w_2, ... w_k$ with the XAI tokens. We define the \textit{'Weighted Explanation Token (WET)'} as the weighted-average derived from the adjusted tokens of the XAI methods. Additionally, the \textit{'High Resolution Token (HRT)'} represents the high-resolution space post the 'non-linear structure' layer in the explanation optimization process, while the \textit{'Low Resolution Token (LRT)'} denotes the input token's low-resolution space.
\par The proposed generalized framework illustrated in Fig. \ref{fig2g} can be applied to enhance explanations across domains such as large language models, multi-modality frameworks, video prepossessing, computer vision models, vision large models, foundation models, etc. with minor modifications to the non-linear structure of the encoder-decoder architecture to suit the specific task at hand.
\section{Discussion}\label{sec12}
The importance of explainability, particularly in critical sectors like healthcare, medicine and the geosciences cannot be overstated. The opacity of deep learning models poses challenges in comprehending their decision-making processes, driving the need for robust methodologies for explainability. However, the high inter-method variability observed in explanations generated by existing XAI methods increase uncertainty and raises questions regarding the true explanation of deep learning networks in computer vision tasks. On top of this, it has been shown that no existing method is optimal across different quality metrics and for various prediction tasks and settings. Hence, there is a necessity for exploring ways to derive unique and optimal explanations building upon existing post-hoc XAI methods and guided by user-defined evaluation metrics. The uniqueness of the derived explanation directly address the issue of the current inter-method variability, while at the same time ensuring explanation optimality. 
\par Our study explores the degree to which such an explanation is even derivable. Our proposed framework, inspired by Herbert Simon’s seminal work on "bounded rationality" and "satisficing" decision-making, is shown to be able to derive such a unique and optimal explanation for pre-trained deep learning networks, tailored to specific computer vision tasks. By integrating diverse explanations from established XAI methods, employing a non-linear network architecture and an innovative loss function, we achieve to reconstruct a unique explanation that faithfully captures the essence of the network's decision-making process and minimizes complexity. To validate our framework, we conducted experiments on multi-class and binary classification tasks in both two- and three-dimensional spaces, focusing on automobiles-animals and neuroscience imaging domains.
\par Our results demonstrate the effectiveness of the framework in achieving high faithfulness and low complexity in explanations. Statistical analysis revealed significant differences between our approach and alternative state-of-the-art methods, further affirming its efficacy. Visual explanations provided by our framework offer insights into the decision-making process of deep learning models. In two-dimensional scenarios, our approach consistently outperformed other XAI methods, emphasizing significant features crucial for classification. In three-dimensional applications, our framework showcased less complexity and focused on specific areas, advancing our understanding of underlying mechanisms. Our study also discusses the rationale behind favoring non-linear optimizing structures over linear alternatives. Importantly, our results highlight that a straightforward linear combination of existing XAI methods does not improve faithfulness or reduce complexity meaningfully, as demonstrated in Figure 5 ('Weighted Average' and 'Explanation Optimizer (linear)'). In contrast, our non-linear explanation optimizer achieves significant advancements, reinforcing the non-trivial nature of designing such a unifying framework.
Our optimizer is successful in part because of the utilization of a unique cost function that consists of explanability scores and a similarity score that constraints the convergence of the optimizer and avoids reaching non meaningful, trivial solutions during training. The generalized version of the proposed framework can be used in any explanation task of a multi-dimensional space. 
\par In terms of future research, the current framework could be expanded to include more explainability metrics in the loss function, beyond faithfulness and complexity that were used here as a first step, for example, robustness, localization etc. Moreover, one could explore whether similar explanation-optimizing frameworks can be derived for global explanation settings. Last, future work should focus on expanding the current framework in including layers that allow the quantification of uncertainty of the explanations. Such an addition shall further assist in drawing robust conclusions about the decision-making process of deep networks. 
\section{Conclusions}
The sub-optimal results of individual XAI methods, combined with high inter-method variability, present a significant challenge in the field of XAI, as this variability undermines trust in deep network predictions. This study addresses this challenge by introducing a novel framework designed to enhance explainability through the integration of multiple established XAI methods. Central to our approach is a non-linear neural network model, the ’explanation optimizer,’ which synthesizes a singular, optimal explanation from these methods. Crucially, this process goes beyond a simple combination or ensembling of existing XAI tools. Instead, it employs a carefully designed nonlinear, constrained optimization technique, leveraging a non-trivial loss function to achieve results that outperform those produced by the individual methods or their straightforward combinations.
\\
The optimizer operates with two key metrics that evaluate the quality of the generated explanations: faithfulness, which measures how accurately the explanation mirrors the network’s decision-making, and complexity, which assesses the comprehensibility of the explanation for end users. By carefully balancing these metrics, the optimizer generates explanations that are both accurate and accessible. Our experimental validation across multi-class and binary classification tasks in both 2D object and 3D neuroscience imaging demonstrates the efficacy of this approach. Notably, our explanation optimizer achieved superior faithfulness scores, with an average increase of 155\% and 63\% over the best-performing individual XAI methods in 3D and 2D applications, respectively, while also reducing complexity to enhance comprehensibility.
\\
This framework represents a significant advancement in XAI by systematically enhancing the explainability of deep learning models. By generating explanations that are both highly faithful and easy to interpret, our approach addresses the long-standing issue of inter-method variability, strengthens trust and interpretability, and promotes the broader adoption of AI technologies in critical fields such as healthcare and medicine.
\clearpage %%Remove this from your manuscript
\section*{Declaration of competing interest}
The authors declare that they have no known competing financial interests or personal relationships that could have appeared to
influence the work reported in this paper.
\section*{Acknowledgment}
All research at the Department of Psychiatry in the University of Cambridge is supported by the NIHR Cambridge Biomedical Research Centre (NIHR203312) and the NIHR Applied Research Collaboration East of England. The views expressed are those of the author(s) and not necessarily those of the NIHR or the Department of Health and Social Care. We acknowledge the use of the facilities of the Research Computing Services (RCS) of University of Cambridge, UK. We acknowledge the South-Eastern Norway Regional Health Authority and the Research Council of Norway (223273). GKM consults for ieso digital health. All other authors declare that they have no competing interests. 
\printcredits
\bibliographystyle{cas-model2-names}
\bibliography{sn-bibliography}
\end{document}